\documentclass[journal,twocolumn]{IEEEtran}
\usepackage{graphicx}
\usepackage{caption}
\usepackage{amsfonts}
\usepackage{subcaption}
\usepackage{amsmath}
\usepackage[square,sort,comma,numbers]{natbib}
\usepackage{booktabs}
\usepackage{url}
\usepackage{algorithm}
\usepackage{algorithmic}
\usepackage{amsmath,amssymb,theorem}
\usepackage{color,soul}
\usepackage{amstext} 
\usepackage{array}   
\newcolumntype{C}{>{$}c<{$}}

\begin{document}
	
	\author{Subhasis~Banerjee~\IEEEmembership{Member,~IEEE},%
		~Sushmita~Mitra,~\IEEEmembership{Fellow,~IEEE,}%
		~Anmol~Sharma$^\ddag$,
		~and~B.~Uma~Shankar$^\dagger$~\IEEEmembership{Member,~IEEE}
		\thanks{$^\dagger$Corresponding author}
		\thanks{S. Banerjee, S. Mitra and B. Uma Shankar are with Machine Intelligence Unit, Indian Statistical Institute, 203 B. T Road, Kolkata 700108, India.
			\emph{E-mail:} mail.sb88@gmail.com, \{sushmita,uma\}@isical.ac.in}
		\thanks{S. Banerjee, acknowledges financial
			support from the Visvesvaraya PhD Scheme by Department
			Of Electronics \& Information Technology (DeitY) Ministry
			of Communications and IT, Government of India}
		\thanks{$^\ddag$Anmol Sharma was a student at the Department of Information Technology, DAV Institute of Engineering and Technology, Kabir Nagar, Jalandhar, India, and Research Trainee at Machine Intelligence Unit, Indian Statistical Institute, 203 B. T Road, Kolkata 700108, India.	\emph{E-mail}: anmol.sharma293@gmail.com}
		\thanks{Anmol Sharma, acknowledges partially supported from Indian National Academy of Engineering (INAE) under the scheme ``Mentoring of Engineering Students by INAE Fellows",
			and this work was carried out at Machine Intelligence Unit, Indian Statistical Institute.}
	}%

	\title{A CADe System for Gliomas in Brain MRI using Convolutional Neural Networks}
	
	\maketitle
	\thispagestyle{empty}
	\begin{abstract}
		Inspired by the  success of Convolutional Neural Networks (CNN), we develop  a novel Computer Aided Detection (CADe) system using CNN for Glioblastoma Multiforme (GBM) detection and segmentation from multi channel MRI data. A two-stage approach first identifies  the presence of GBM. This is  followed by a GBM localization in each  ``abnormal'' MR slice. As part of the CADe system,  two CNN architectures viz. Classification CNN ({\scshape{c-cnn}}) and Detection CNN ({\scshape{d-cnn}}) are employed.  The CADe system considers MRI data  consisting of four sequences ($T_1$, $T_{1c}$, $T_2$, and $T_{2FLAIR}$) as input, and  automatically generates the bounding boxes encompassing the tumor regions in each slice which is deemed abnormal. Experimental results demonstrate that the  proposed CADe system, when used as a preliminary step before segmentation, can allow improved delineation of tumor region while reducing false positives arising in normal areas of the brain. The  GrowCut  method, employed  for tumor segmentation,  typically requires a foreground and background seed region for initialization. Here  the algorithm is initialized with seeds automatically generated from  the output of the  proposed CADe system, thereby resulting in improved performance as  compared to that using  random seeds.
	\end{abstract}

\begin{IEEEkeywords}
	\noindent Convolutional neural network, deep learning, gliomas, MRI, brain tumor segmentation, bounding box, CADe.
\end{IEEEkeywords}

\section{Introduction}

Brain tumors are one of the deadliest cancers with a high mortality rate \cite{DeAngelis2001, Bauer2013}. They can be primary, i.e. directly originating in the brain, or metastatic, i.e. spreading from other parts of the body. Gliomas constitute 70\% of malignant primary brain tumors in adults \cite{Bauer2013}, and are usually classified as High Grade Gliomas (HGG) and Low Grade Gliomas (LGG). The HGG encompasses grades III and IV of the WHO categorization \cite{Louis2007_WHO},  exhibiting a rapidly proliferating behaviour with a patient survival period of only about a year \cite{Bauer2013}.

Magnetic Resonance Imaging (MRI), Positron Emission Tomography (PET) and Computed Tomography (CT) are some of the standard radio imaging techniques used for diagnosing abnormalities in the brain.  MRI has been extensively employed in diagnosing brain and  nervous system abnormalities, over the last few decades, due to its improved soft tissue contrast as compared to plain radiography or CT \cite{Edelman1993_MRI,mitra2014integrating,mitra2015medical}. MR images are usually procured in multiple sequences or modalities, depending on the different excitation and repetition times used during the scan. This enables the  capture of  distinct structures of interest, by producing noticeably different tissue contrasts \cite{Bauer2013,banerjee2016novel}.
The sequences include $T_1$-weighted, $T_2$-weighted, $T_1$-weighted with contrast enhanced ($T_{1c}$), and $T_2$-weighed with fluid-attenuated inversion recovery ($T_{2FLAIR}$). The rationale behind using these four sequences lies in the fact that different tumor regions may be visible in different sequences, allowing for a more accurate composite marking of the  tumor region. Delineation of tumor region in MRI sequences is of great importance since it allows: \textit{i)} volumetric measurement of the tumor, \textit{ii)} monitoring of tumor growth in the patient between multiple MRI scans, and \textit{iii)} treatment planning with follow-up evaluation.

Tumor segmentation from brain MRI sequences is usually done manually by the radiologist. Being a highly tedious and error prone task, mainly due to factors such as human fatigue, overabundance of MRI slices per patient, and increasing number of patients, manual operations often lead to inaccurate delineation. Moreover, use of qualitative measures of evaluation by radiologists results in  high inter- and intra-observer error rates, which are often difficult to characterize \citep{Bauer2013,banerjee2016single,banerjee2017roi,Pereira2016}. The need for an automated or semi-automated Computer Aided Detection/Diagnosis (CADe/CADx) system thus becomes apparent. Such a system improves the overall performance of the detection and subsequent segmentation of abnormalities, particularly when used as an assistant to the radiologist.

Automated detection is a challenging task due to the variety of shapes, textures and orientations exhibited by the tumor region. Typically the  tumor acts as a mass and pushes the normal tissue, thereby changing the overall structure of the brain. Besides, brain MRI slices are known to be affected by Bias Field Distortion (BSD) and other artifacts that change the homogeneity of tissue intensities in different slices of the same brain. Existing methods leave significant room for increased automation, applicability and improved accuracy. Since segmentation of tumors in brain is typically preceded by its detection and plays an important role in curbing improper segmentation of tumor region, we investigate here the automated tumor detection problem in brain MR images.

Recently, deep learning research has witnessed a growing interest for data analysis. Deep learning is a branch of machine learning consisting of a set of algorithms that attempt to model high level abstractions in data by using a deep model with multiple processing layers, composed of both linear and non-linear transformations \cite{LeCun2015,Bengio2016_book,Lecun1998}. Among these,  Convolutional Neural Networks (CNNs) provided  impressive performance on image recognition and classification problems \cite{Krizhevsky2012, Farabet2013, Zhang2014_overfeat, Simonyan2014_verydeepcnn}.

Convolutional Neural Networks (also called ConvNets or CNNs) \cite{Lecun1998} are  suitable for processing input that comes in the form of a grid-like topology, for instance -- time series and image data. Unlike  a traditional Artificial Neural Network (ANN), a CNN uses a convolution operation instead of  matrix multiplication in some or all of its layers. The design of  CNNs  is motivated by the functioning of the mammalian vision system, which hierarchically captures semantically rich visual features \cite{Hubel1962, Fukushima1980, Lecun1998}. User-provided bounding boxes are a simple and popular form of annotation used in computer vision to initialize object segmentation as well as induce spatial constraints. DeepCut \cite{Deepcut-2017} combines CNN with  iterative  graphical optimization  to recover pixelwise object segmentations, from an image database with existing bounding box annotation. Bounding  boxes are manually generated from user-provided segmentation. A fully connected conditional random field serves to regularize the segmentation. Experimental results demonstrate segmentation of the brain and lung of fetal MRI. However such manual annotation entails human bias, is prone to error, and is also time consuming.

Our research focuses on the design and development of a  fully automated CADe System for the detection of HGG using CNNs. The novel CADe system first identifies the presence of a tumor from the 3D MR slices of the brain. The bounding box approach automatically localizes the tumor in each ``abnormal'' slice, encompassing $T_1, T_{1C}, T_2, T_{2FLAIR}$ sequences. Subsequent segmentation enables improved tumor delineation, with reduced false positives. Initial seeds for segmentation are automatically generated by the system.

The rest  of the paper is organized as follows.  Section \ref{sec:previous_work} provides a brief literature review  on detection of brain tumors.  Section \ref{sec:contrib} highlights the characteristics and merits of the proposed CADe  system, while outlining the architecture and methodology.   Section \ref{sec:setup} describes the experimental results  on the BRATS 2015 dataset, demonstrating the effectiveness of subsequent segmentation both qualitatively and quantitatively with respect to existing related methods. Finally conclusions are presented in Section \ref{sec:conclusion}.

\section{Overview of Brain Tumor Detection}
\label{sec:previous_work}
Over the years, a number of techniques have been successfully devised to automatically detect brain tumors. Generative model based approaches define \emph{a-priori} model of the normal brain,  and detect abnormal regions by looking for outliers \cite{Prastawa2004_gen, Cuadra2004_atlas, Zacharaki2008_gen, Menze2010_gen}. Other generative models may use asymmetry cues to identify abnormalities in the brain MRI, with the underlying assumption that the left and right halves of the brain are symmetric for a normal patient \cite{Khotanlou2009_3d, Saha2012_quick}.
Saha \emph{et al.}  \cite{Saha2012_quick} employed the concept of bounding boxes for glioma and edema detection from brain MRI slices. The method uses symmetry based \emph{a-priori} assumptions, with  the left and right halves of the brain being expected to be a asymmetric in case of possible tumors. A  scoring function based on Bhattacharya coefficient is computed with gray level intensity histograms. Although generative models have been shown to generalize well on unseen data due to their simple hypothesis functions, yet their dependence on\emph{ a-priori} knowledge makes them unsuitable to applications where this is not available. Moreover, these models  heavily rely on accurate registration for aligning images of different modalities; which is sometimes problematic in the presence of an abnormality in the brain \cite{Parisot2012_joint}.  Some of the ``atlas'' based methods \cite{Cuadra2004_atlas} may also lead to incorrect learning in  the presence of  large deformations in  brain structures \cite{Khotanlou2009_3d}.

Image processing based methods, on the other hand,  perform various operations on the MRI slices to detect abnormal (tumor) regions. They exploit underlying differences in intensity values between normal and abnormal regions. This encompasses  watershed segmentation \cite{Mustaqeem2012efficient} followed by the application of some morphological operations to detect tumor regions in an MRI slice \cite{banerjee2016single,Sharma2014brain}. However  image processing based approaches often suffer from severe over-segmentation and noise, in the form of false positive regions, resulting in inappropriately delineated tumor region.

Advances in machine learning have made an impact over research in brain tumor detection from  MRI slices. Most of the literature in this field proposed the use of hand-crafted features such as fractals \cite{Islam2013multifractal}, Gabor coefficients \cite{Subbanna2013, Wu2014_des}, or their combination  \cite{Soltaninejad2016}. These features are then used to train  AdaBoost \cite{Islam2013multifractal}, Bayesian classifier \cite{Subbanna2013}, decision trees, forests and SVMs \cite{Soltaninejad2016,Zikic2012decision} which then detect and delineate the tumor region in the MRI slice(s). Although the above approaches demonstrate good performance on BRATS datasets, they  rely heavily on hand-crafted features  requiring extensive domain knowledge of the data source. Manual design of  features typically demands greater insight into the exact characteristics of normal and abnormal tissues in the brain. Moreover, such features may not  be  able to accurately capture the important representative features in the abnormal tumor regions of the brain; leading to hindrance in classifier performance.

CNNs essentially revolutionized the field of computer vision and have since become the de-facto standard for various object detection and recognition tasks \cite{Farabet2013, Goodfellow2013_captcha, Zhang2014_overfeat, Simonyan2014_verydeepcnn}. These networks automatically learn mid-level and high-level representations or abstractions from the input training data in the form convolution filters, that get updated during the training process. They work directly on raw input (image) data, and learn the underlying representative features of hierarchically complex  input, thereby ruling out the need for specialized hand-crafted image features. Moreover, CNNs  require no prior domain knowledge and can learn to perform any task  by automatically working through the training data.

A CNN  is built using two fundamental types of layers, namely the Convolution Layer and Pooling Layer.
The inputs percolating through network are the responses of convoluting the images with various filters. These filters act as detectors of simple patterns like lines, edges, corners, from spatially  contiguous regions in an image. When arranged in many layers, the filters can automatically detect prevalent patterns while blocking irrelevant regions. The pooling layers serve to down sample the convoluted response maps. This helps lessen  the number of  trainable parameters, thereby resulting in reduction of overfitting possibilities. Deeper layers help the CNN extract higher levels of feature abstractions.
These layers are usually followed by a classifier, which in most cases is a multi-layer perceptron (MLP). Apart from connection weights inside the MLP, the other trainable parameters in a CNN  are the filters in each convolution layer.

The adoption rate of CNNs in medical imaging  has been on the rise \cite{Greenspan2016_guest_survey}, with recent research focusing on topics ranging from lesion detection \cite{Setio2016pulmonary, Summers2015improving, Dou2016automatic, Sirinukunwattana2016locality} to segmentation and shape modelling \cite{Ghesu2016marginal, Brosch2016deep, Pereira2016} from 2D/3D CT and MR images. Inspired by their success, many  medical imaging researchers   have applied CNNs as pixel-level classifiers for abnormality detection and segmentation in brain MRI. Urban \emph{et al.} \cite{Urban2014_brats} used 3D CNNs for detecting abnormal voxels from volumetric MRI sequences. Havaei \emph{et al.} \cite{Davy2015_brats} designed  a 2-way CNN architecture that exploits both the local and global context of an input image. Each pixel in every 2D slice of the MRI data is classified into either normal or a part of the tumor region. Recently, Pereira \emph{et al.} \cite{Pereira2016} demonstrated impressive results by developing  two separate CNN  architectures corresponding  to  pixel-wise label prediction  for detecting HGG and LGG tumor regions in brain MRI slices.

However existing literature  using CNNs mainly focuses on pixel (or voxel) level tumor detection by labelling normal or abnormal categories \cite{Urban2014_brats,Zikic2014_brats,Davy2015_brats, Pereira2016}.
In this process, the two distinct phases  \textit{detection} and \textit{segmentation} of tumor regions get merged. Although this might appear to be  an ideal scenario, where   the detection phase get bypassed, yet  this  may lead to high false positive rates because  the algorithm works on every pixel in the MRI slice and is not constrained inside a specific  region. Even in clinical settings, the demarcation between a normal and an abnormal patient followed by the detection of an abnormal region assumes greater  significance; and this always precedes the actual segmentation and volumetric analysis of the tumor region. The ever-increasing deluge  of data, that the radiologists are regularly  besieged with, becomes  a major hindrance towards the accurate delineation; thereby highlighting  the need for an automated detection system.

The premise of this paper is that optimal tumor segmentation can be achieved through a preceding approximate tumor detection or localization step, that can aid accurate segmentation by acting as a seed towards constrained segmentation. Hence we take a detection-first approach in which the tumor region is approximately detected by our proposed system. Next this information  is used to generate the seed for segmentation, resulting in the whole tumor region getting accurately  delineated.

\section{The CADe System}
\label{sec:contrib}

A novel  Computer Aided Detection (CADe) system is designed  for tumors in Brain MRI slices, employing  a combination of classification and regression phases. A pair of convolution network  architectures, viz. {\scshape{c-cnn}} and {\scshape{d-cnn}}, constitute the CADe system. At the entry point to the system, the Classification Convolutional Neural Network ({\scshape{c-cnn}}) determines whether  or not the patient's brain MRI study is normal (or abnormal) based on the presence (or absence) of suspicious regions. Once an abnormal sample  is identified, the Detection Convolutional Neural Network ({\scshape{d-cnn}}) is invoked to approximately identify the abnormal regions in each MRI slice. {\scshape{d-cnn}} works by predicting a bounding box around the tumor region to identify the abnormality.

\begin{figure}[]
\begin{center}
	\includegraphics[scale = 0.5]{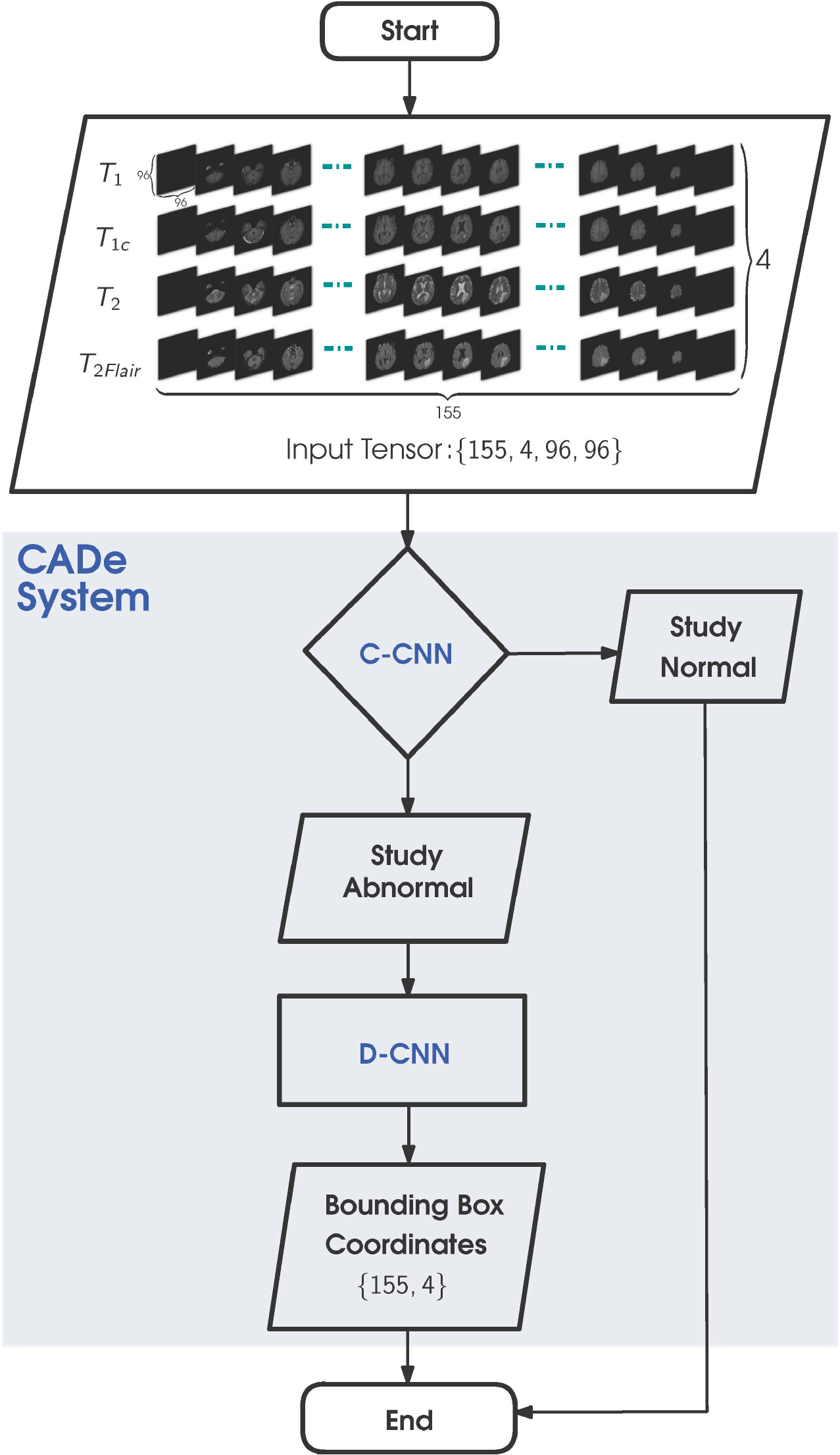}
	\caption{Flowchart illustrating the CADe system for brain MRI}
	\label{fig:cade}
\end{center}
\end{figure}

We tackle the problem of tumor detection in brain MRI using a bounding box based localization approach, as evident in  the computer vision community. The proposed method is robust to any anatomical changes in the appearance of the brain, as well as towards improper registration of MRI slices. The schematic diagram of the CADe system is provided  in Fig. \ref{fig:cade}. The input to the system is a patient study containing 4-sequence $(96 \times 96)$ MRI slices, and output is an approximate localization of any abnormality in the slices in the form of bounding box coordinates. When used  as a preceding  step to tumor  segmentation, it can provide a seed for constrained  demarcation  of the abnormal region; thereby  leading to improved delineation of the tumor region while simultaneously reducing the number of false positives.   The approximate tumor position predicted by the  CADe system is used as  seed for GrowCut \cite{Vezhnevets2005_growcut} towards subsequent  segmentation of the  tumor region from the MRI slice.

\subsection{Contribution}

The merits of our  CADe system,  over existing  tumor detection methodologies for brain MRI, are outlined below.

\begin{itemize}
	\item Due to the discriminative nature of our  CADe system, there is neither any requirement of inherent \emph{a-priori} domain knowledge nor assumption of brain symmetry (as in generative models \cite{Prastawa2004_gen, Cuadra2004_atlas, Zacharaki2008_gen, Menze2010_gen}). The deterministic nature of the system also rules out any inter-observer error, as  is prevalent in clinical setting.
	
	\item Compared to earlier machine learning based models \cite{Zikic2012decision, Islam2013multifractal, Subbanna2013, Wu2014_des,Soltaninejad2016}, our system eliminates the need of  hand-crafted features for slice classification and tumor localization by automatically extracting / learning the underlying highly representational and hierarchical features.
	
	\item Due to the preceding approximate localization step, the final tumor segmentation can be constrained to the specific suspicious region(s); thereby ruling out any false positives in other (normal) regions of the brain.
	
	\item Unlike   atlas based approaches \cite{Khotanlou2009_3d}, the proposed system is highly robust to significant changes and deformations in brain anatomy casued by the  presence of abnormality (tumor).
	
\end{itemize}

\subsection{Preprocessing}
MRI sequence slices usually suffer from inconsistent image intensity problem, better known as Bias Field Distortion (BFD). This makes the intensity of the same tissue to vary across different slices of a sequence for a single patient. Thus the input training data is first subjected to Bias Field Distortion correction using N4ITK \cite{Tustison2010_n4itk}  for a homogeneous intensity range throughout each sequence. Further, the images are processed with a median filter to rule out any high frequency image noise. The images in both training and testing sets are standardized to zero mean and unit variance by calculating mean intensity value and standard deviation of pixels in the training set.

\subsection{Proposed architecture}\label{baseline}
The two-stage architecture, consisting of the classifier and detection modules {\scshape{c-cnn}} and {\scshape{d-cnn}}, serves to classify a 2D brain MRI slice into normal (or abnormal) followed by an approximate localization of the tumor region in the specified slices. This is outlined in Fig.~\ref{fig:cade}.

\subsubsection{Classification ConvNet ({\scshape{c-cnn}})}
\begin{figure*}[]
\begin{center}
	\includegraphics[scale = 1, width = 16cm]{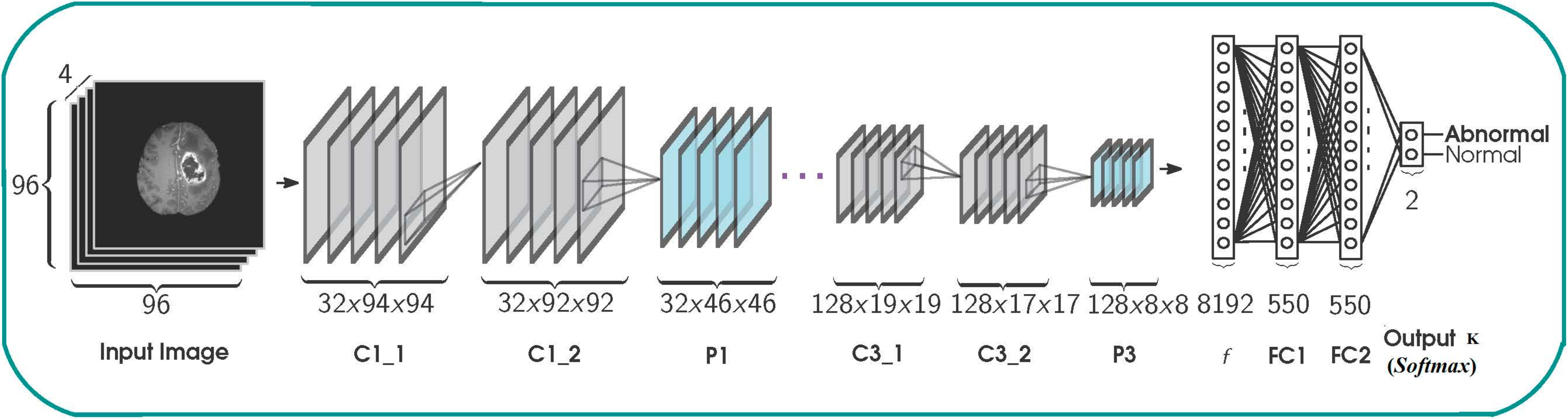}
	\caption{Network {\scshape{c-cnn}}}
	\label{fig:c-cnn}
\end{center}
\end{figure*}

A 12 layer Classification network {\scshape{c-cnn}}, consisting of three sets of stacked convolution and pooling layers followed by two fully connected layers, is  illustrated in Fig. \ref{fig:c-cnn}. This network serves as the entry point of the CADe system, which takes each 2D brain MRI slice $I \in \mathbb{R}^{4 \times 96 \times 96}$ (four-sequence MRI slice of size $96 \times 96$) as input and provides the probability of that slice being normal or abnormal as output. The network thus classifies each slice, and computes  the overall number of slices being flagged as abnormal in a particular study. If more than 5\% of the slices are flagged as abnormal, the patient study is then passed down the pipeline to the  localization network {\scshape{d-cnn}}. The value 5\% was chosen empirically using a small validation set.

The {\scshape{c-cnn}} network consists of six convolution layers (C1\_1, C1\_2, C2\_1, C2\_2, C3\_1, C3\_2), with  filter (or kernel) sizes $3 \times 3$ but having increasing filter numbers ($32, 64, 128$) over the layers. There are three  pooling layers P1, P2, P3 with filter size $ 2 \times 2$ each.  The classifier at the end is a fully connected MLP of connectivity $8192 \times 550 \times 550 \times 2$. Table \ref{tab:c-cnn} summarizes the entire {\scshape{c-cnn}} architecture. Smaller kernels produce  better regularization  due to the smaller number of trainable weights,  with the possibility  of constructing  deeper networks  without losing too much information in the layers \cite{Pereira2016, Simonyan2014_verydeepcnn}. Greater number of  filters, involving deeper  convolution layers, allows for  more feature maps to be generated; thereby  compensating for the decrease in the size of each feature map caused by  ``valid'' convolution and pooling layers. The convolution layer is said  to be of type  ``valid'' when the input to the layer is not zero-padded before the convolution operation,  such that  the resulting output becomes gradually smaller  down the layers in terms  of input  dimension.

\begin{table}
	\setlength\tabcolsep{1.2mm}
	\centering
	\caption{Architecture configuration of {\scshape{c-cnn}} for 2D MRI slice classification}
	\begin{tabular}{lcccccc}
		\toprule
		\textbf{Layer} &  \textbf{Filter} & \textbf{Stride} & \textbf{FC} & \textbf{Conv} & \textbf{Input} & \textbf{Output}  \\
		&  \textbf{Size} &  & \textbf{Units} & \textbf{Type}& &   \\
		\midrule
		C1\_1 & 32\text{x}3\text{x}3   & 1   &    --   & \text{valid}  & 4\text{x}96\text{x}96 & 32\text{x}94\text{x}94  \\
		C1\_2 & 32\text{x}3\text{x}3   & 1   &    --   & \text{valid}  & 32\text{x}94\text{x}94 & 32\text{x}92\text{x}92  \\
		P1 &    2\text{x}2             & 2   &   --    &   --    & 32\text{x}92\text{x}92 & 32\text{x}46\text{x}46  \\
		C2\_1 & 64\text{x}3\text{x}3   & 1   &    --   & \text{valid}  & 32\text{x}46\text{x}46 & 64\text{x}44\text{x}44  \\
		C2\_2 & 64\text{x}3\text{x}3   & 1   &    --   & \text{valid}  & 64\text{x}44\text{x}44 & 64\text{x}42\text{x}42  \\
		P2 &    2\text{x}2             & 2   &   --    &   --    & 64\text{x}42\text{x}42 & 64\text{x}21\text{x}21  \\
		C3\_1 & 128\text{x}3\text{x}3  & 1   &    --   & \text{valid}  & 64\text{x}21\text{x}21 & 128\text{x}19\text{x}19  \\
		C3\_2 & 128\text{x}3\text{x}3  & 1   &    --   & \text{valid}  & 128\text{x}19\text{x}19 & 128\text{x}17\text{x}17  \\
		P3 &    2\text{x}2             & 2   &   --    &   --    & 128\text{x}17\text{x}17 & 128\text{x}8\text{x}8  \\
		
		FC1 &      --    &    --   & 550  &   --    & 8192  & 550   \\
		FC2 &     --    &    --   & 550  &  --     & 550  & 550   \\
		Out (K) &      --    &   --    & 2     &     --  & 550  & 2      \\
		\bottomrule
	\end{tabular}%
	\label{tab:c-cnn}%
\end{table}%

The output feature map dimension, from a convolution layer, is  calculated as
\begin{equation}
w_{out}/h_{out} = \frac{(w_{in}/h_{in} - F + 2P^{\prime})}{Stride} + 1,
	\label{eq:getSize}
\end{equation}where $w_{in}$ is the input image width, $h_{in}$ is input image height, $w_{out}$ is effective output width, and $h_{out}$ is output height. Here $P^{\prime}$ denotes  the input padding which (in our case) is set to zero due to ``valid" convolution  involving nil  zero-padding. The displacement   $Stride=1$, with $F$ being the receptive field  (kernel size) of the neurons in a particular layer.
Input downsizing  in max pooling layer, with filter size fixed at  $2 \times 2$ and stride of two for a non-overlapping pooling operation,  results in  downsampling  by a factor of $2$.

The final feature maps from the layer P3 are  flattened into a feature vector $f \in \mathbb{R}^{8192}$, before being fed to the fully connected  layer FC1 of the classifier MLP.
Two fully connected layers, with $550$   hidden neurons each, constitute the MLP having two final outputs.  The number of hidden neurons are chosen through automatic hyperparameter estimation using cross-validation. Non-linearity in the form of Rectified Linear Unit (ReLU) \cite{Nair2010_relu}  is applied after each convolution as well as  fully connected layer, thereby  transforming negative activation values $a$  to zero using $\max(0, a)$. Finally, the predicted distribution $S(a)$ is computed by taking the softmax

\begin{equation}
 S(a)\:=\:\frac{e^{a_j}}{\sum^{K}_{k=1}e^{a_k}},
\end{equation} where $K = 2$  corresponds to the number of output neurons  and $a_{k}$ is the  activation value of  $k$th neuron. The number of trainable parameters in {\scshape{c-cnn}} is 5,097,598.

\begin{figure*}
\begin{center}
	\includegraphics[scale = 1, width = 16cm]{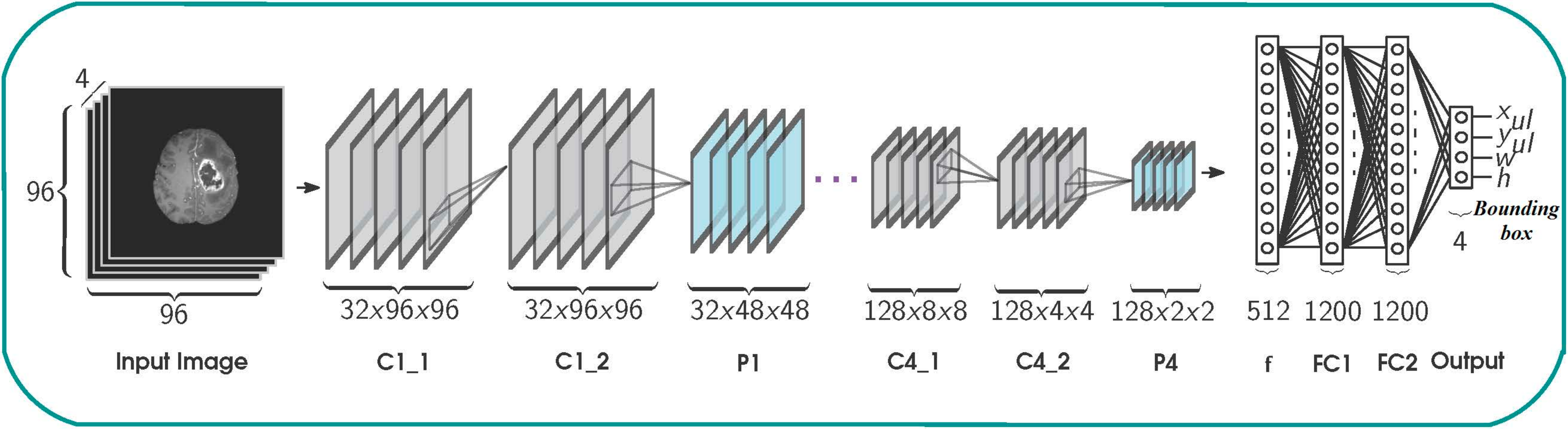}
	\caption{Network {\scshape{d-cnn}}}
	\label{fig:d-cnn}
\end{center}
\end{figure*}

\subsubsection{Detection ConvNet ({\scshape{d-cnn}})} The $15$ layer Detection network {\scshape{d-cnn}},  for predicting an approximate bounding box around the tumor region, is depicted in Fig.~\ref{fig:d-cnn}.   Its  input is a 4-sequence brain MRI slice $I \in \mathbb{R}^{4 \times 96 \times 96}$,  and its  output consists of  four real numbers $y = \left\lbrace x_{ul}, y_{ul}, width, height\right\rbrace $,  where $x_{ul}$, $y_{ul}$ are the abscissa and ordinate of the  upper left corner of the bounding rectangle, respectively,  with  $width$  and $height$  referring  to its  corresponding dimensions. {\scshape{d-cnn}} consists of four  sets of stacked convolution and pooling layers, followed by two fully connected layers. Due to the complexity of the  bounding box prediction problem, the {\scshape{d-cnn}} network architecture is deeper as compared to the  {\scshape{c-cnn}}. Table \ref{tab:d-cnn}  summarizes the entire {\scshape{d-cnn}} architecture.

Convolution layers in {\scshape{d-cnn}} have  filter numbers ($32, 64, 128, 128$), while the filter sizes are $3 \times 3$  in the first three pairs of layers and  $5 \times 5$ in the last layer. The convolution type in  the first three layers of the {\scshape{d-cnn}} are set to ``same"  (allowing input zero-padding to preserve spatial size),  with  the last pair of  layers  being of type ``valid"  involving no  zero-padding at input. The feature map generated after  layer P4 is  flattened into a feature vector  $f \in \mathbb{R}^{512}$, and  fed into the first fully connected layer FC1 with $1200$ hidden neurons (chosen through automatic cross validation).  As in  {\scshape{c-cnn}}, the non-linearities after each convolution layer are set to ReLU; although  no such non-linearity is applied after the last output layer. Note that the total number of trainable parameters in {\scshape{d-cnn}} is  2,760,612, which is  reduced as compared to that of the {\scshape{c-cnn}}.

\begin{table}
	\setlength\tabcolsep{1.2mm}
	\centering
	\caption{Architectural configuration of {\scshape{d-cnn}} for approximate tumor localization}
	\begin{tabular}{lcccccc}
		\toprule
		\textbf{Layer} &  \textbf{Filter} & \textbf{Stride} & \textbf{FC} & \textbf{Conv} & \textbf{Input} & \textbf{Output}  \\
		&  \textbf{Size} &  & \textbf{Units} & \textbf{Type}& &   \\
		\midrule
		C1\_1 & 32\text{x}3\text{x}3   & 1   &    --   & \text{same}  & 4\text{x}96\text{x}96 & 32\text{x}96\text{x}96  \\
		C1\_2 & 32\text{x}3\text{x}3   & 1   &    --   & \text{same}  & 32\text{x}96\text{x}96 & 32\text{x}96\text{x}96  \\
		P1 &    2\text{x}2   & 2   &   --    &   --    & 32\text{x}96\text{x}96 & 32\text{x}48\text{x}48  \\
		
		C2\_1 & 64\text{x}3\text{x}3   & 1   &    --   & \text{same}  & 32\text{x}48\text{x}48 & 32\text{x}48\text{x}48  \\
		C2\_2 & 64\text{x}3\text{x}3   & 1   &    --   & \text{same}  & 32\text{x}48\text{x}48 & 32\text{x}48\text{x}48  \\
		P2 &    2\text{x}2   & 2  &   --    &   --    & 32\text{x}48\text{x}48 & 32\text{x}24\text{x}24  \\
		
		C3\_1 & 128\text{x}3\text{x}3   & 1   &    --   & \text{same}  & 64\text{x}24\text{x}24 & 64\text{x}24\text{x}24  \\
		C3\_2 & 128\text{x}3\text{x}3   & 1   &    --   & \text{same}  & 64\text{x}24\text{x}24 & 64\text{x}24\text{x}24  \\
		P3 &    2\text{x}2   & 2  &   --    &   --    & 64\text{x}24\text{x}24 & 64\text{x}12\text{x}12  \\
		
		C4\_1 & 128\text{x}5\text{x}5   & 1   &    --   & \text{valid}  & 64\text{x}12\text{x}12 & 128\text{x}8\text{x}8  \\
		C4\_2 & 128\text{x}5\text{x}5   & 1   &    --   & \text{valid}  & 128\text{x}8\text{x}8 & 128\text{x}4\text{x}4  \\
		P4 &    2\text{x}2   & 2 &   --    &   --    & 128\text{x}4\text{x}4 & 128\text{x}2\text{x}2  \\
		
		FC1 &      --    &    --   & 1200  &   --    & 512  & 1200   \\
		FC2 &     --    &    --   & 1200  &  --     & 1200  & 1200   \\
		Out &      --    &   --    & 4     &     --  & 1200  & 4      \\
(Bounding Box) &         &     &     &      &   &       \\
		\bottomrule
	\end{tabular}%
	\label{tab:d-cnn}%
\end{table}%

\subsection{Methodology} In this section we briefly describe issues related to parameter selection, cost function, and network evaluation.
\subsubsection{Parameter selection}
The final architecture is  chosen heuristically, with a deep network  developed to overfit followed by regularization using Dropout \cite{Srivastava2014dropout} with a probability $p$. A value of  $p=0.2~(0.5)$ is used in {\scshape{c-cnn}} ({\scshape{d-cnn}}).

\begin{table}
	\centering
	\caption{Hyperparameters  chosen using cross-validation}
	\label{tab:hyper}
	\begin{tabular}{lll}
		\toprule
		\multicolumn{1}{l}{\textbf{Name}}                                      & \multicolumn{1}{l}{\textbf{Hyperparameter}}                                                                                      & \multicolumn{1}{l}{\textbf{Value}}                                                                       \\
		\toprule
		\textbf{Initialization}                                                   & \begin{tabular}[c]{@{}l@{}}weights\\ bias\end{tabular}                                                                   & \begin{tabular}[c]{@{}l@{}}Glorot Uniform (initializer) \cite{Glorot2010_uni}\\ Glorot Uniform (initializer) \cite{Glorot2010_uni}\end{tabular}                          \\
		\midrule
		\begin{tabular}[c]{@{}l@{}}\textbf{Dropout}\end{tabular} & \begin{tabular}[c]{@{}l@{}}$p_{(\mbox{\scshape{c-cnn}})}$\\
			 $p_{(\mbox{\scshape{d-cnn}})}$\end{tabular}                                                        & \begin{tabular}[c]{@{}l@{}}$0.2$\\ $0.5$\end{tabular}                                               \\
			\midrule
		\textbf{Training}                                                         & \begin{tabular}[c]{@{}l@{}}optimizer \\ iterations (\scshape{{\scshape{c-cnn}}})\\ iterations ({\scshape{d-cnn}}) \\ batch\_size \\ learning rate $l_r$\\ $\rho$\\ $\epsilon$\end{tabular} & \begin{tabular}[c]{@{}l@{}}\scshape{AdaDelta}\\ $50$\\ $150$\\ $200$ \\ $1.0$\\ $0.95$  \cite{Zeiler2012_adadelta}\\ $10^{-8}$ \cite{Zeiler2012_adadelta}\end{tabular} \\ \bottomrule
	\end{tabular}
\end{table}

The hyperparameters  required for the training process, provided in  Table \ref{tab:hyper}, were chosen through  automatic cross-validation. While $32,550$ slices were used to train the {\scshape{c-cnn}}, the system had $16,800$ abnormal slices for the {\scshape{d-cnn}}. Since deep CNNs entail  a large number of free trainable parameters, the effective number of training samples were artificially enhanced using real time data augmentation in the form of horizontal and vertical image flipping. This type of augmentation works on the CPU parallel to the training process running on GPU, thereby saving computing time and improving resource usage when the  CPU is idle during training. The weights were updated by  {\scshape{adadelta}} \cite{Zeiler2012_adadelta}  based on Stochastic Gradient Descent (SGD), which adapts the learning rate using first order information. Its  main advantage  lies in  avoiding  manual tuning of learning rate and  is  robust to  noisy gradient values, different model architectures, various data modalities and selection of hyperparameters \cite{Zeiler2012_adadelta}.

\subsubsection{Cost function}
The cost function for {\scshape{c-cnn}} was chosen as  binary cross-entropy (for the two-class problem) as
\begin{equation}
	L_C = \sum_{i=1}^{n}\left\{ -y_i \log(f_i) - (1 - y_i) \log(1 - f_i)\right\},
\end{equation}
where $n$ is the number of samples, $y_i$ is the true label of a sample and $f_i$ is its  predicted label.

In the case of  {\scshape{d-cnn}} the  Mean Squared Error (MSE) was used as the  cost function.

\begin{equation}
L_D = \frac{1}{4n} \sum_{i=1}^{n} \sum_{j=1}^{4}(f_{ij} - y_{ij})^2,
\end{equation}
\noindent where $y_i, f_i$ are vectors with four components corresponding to the four output values.

\subsubsection{Network evaluation}
The {\scshape{c-cnn}}  was evaluated  on the basis of classification accuracy, Area Under the ROC Curve ($AUC$), precision, recall, and $F_{\beta}$ scores. Let $TP$ = true positives, $TN$ = true negatives, $P$ = total number of positive samples, $N$ = total number of negative samples, $FP$ = false positives, and $FN$ = false negatives.
We have
\begin{equation}
\text{Accuracy} = \frac{TP + TN}{P + N},
\label{eq:accuracy}
\end{equation}

\begin{equation}
\text{Precision} = \frac{TP}{TP + FP},
\label{eq:precision}
\end{equation}

\begin{equation}
\text{Recall} = \frac{TP}{TP + FN},
\label{eq:recall}
\end{equation}

\begin{equation}
F_{\beta} = \left( 1 + \beta^2\right) * \frac{precision * recall}{\left( \beta^2 * precision\right) + recall},
\label{eq:beta}
\end{equation}
\noindent with  $\beta$  being chosen as  1 to provide  equal weight to both precision and recall scores.

Evaluation of {\scshape{d-cnn}}, with respect to bounding box detection, was performed using Mean Absolute Error (MAE) and Dice Similarly Co-efficient (DSC). Here
\begin{equation}
\text{MAE} = \frac{1}{n}\sum_{i = 1}^{n} \sum_{j = 1}^{4} |f_{ij} - y_{ij}|
\label{eq:mae}
\end{equation}
 \noindent denotes the number of pixels by which the predicted bounding box is displaced from the original ground truth rectangle, with lower
  values  implying  better prediction.

 A measure of the overlap between the  predicted  and  target bounding boxes was obtained as

\begin{equation}
	\text{DSC} = \frac{2|X \cap Y|}{|X| + |Y|},
\label{eq:dsc}
\end{equation}
where $X, Y$ denote the binary prediction and target masks, respectively.
 The intensity values of masks are either 0 (area outside rectangle) or 1 (area inside rectangle),
   with $0 \leq DSC \leq 1 $  and ``one'' implying a perfect overlap.

\subsection{Segmentation}
The detected tumor region was next segmented by GrowCut \cite{Vezhnevets2005_growcut}, using seeds automatically generated by the
proposed CADe system by Algorithm \ref{alg:genseeds}.

\begin{algorithm}
	\small
	\flushleft
	\caption{Generating  seeds from bounding box}
	\begin{algorithmic}[1]
		\renewcommand{\algorithmicrequire}{\textbf{Input:}}
		\renewcommand{\algorithmicensure}{\textbf{Output:}}
		\REQUIRE $y = \left\lbrace x_{ul}, y_{ul}, width, height\right\rbrace $
		\ENSURE  $seeds =\left\lbrace x_f,y_f, r_f, x_b, y_b, r_b  \right\rbrace$
	
		\STATE $x_f$ = $x_{ul} + width/2$
		\STATE $y_f$ = $y_{ul} + height/2$
		\STATE $r_f$ =  $min(width, height)$ * $0.2$
		
		\STATE $r_b$ = $max(width, height)$ / $2$
		\STATE $x_b$ = $x_f$
		\STATE $y_b$ = $y_f$
		
		\RETURN $\left\lbrace x_f,y_f, r_f, x_b, y_b, r_b  \right\rbrace$
		
	\end{algorithmic}
	\label{alg:genseeds}
\end{algorithm}

\begin{figure}
	\centering
	\includegraphics[width=4cm]{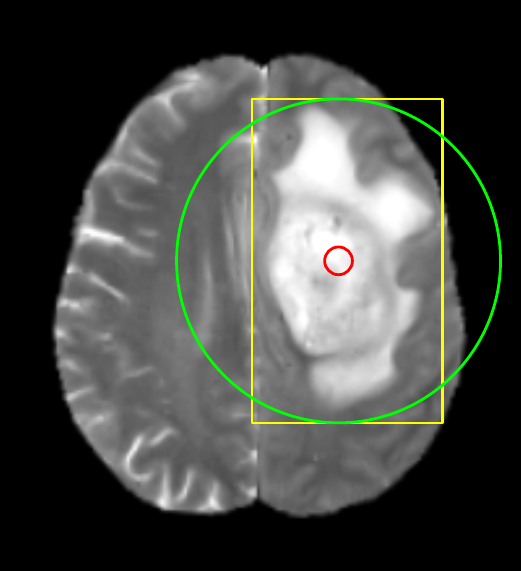}
	\caption{Choice of seeds  by Algorithm \ref{alg:genseeds} on $T_2$ slice. The yellow rectangle denotes predicted bounding box, red circle indicates
 foreground region, and green circle denotes  background region.}
 	\label{fig:seedIllustration}
\end{figure}

The iterative method grows a spline or boundary, inside and outside the bounding box,  to distinguish between the foreground (tumor) and background regions. The ``seed pixels" are chosen  along the circumference of the circular regions having centers ($x_f$, $y_f$) and ($x_b$, $y_b$), and radii $r_f$ and $r_b$,  as depicted in Fig. \ref{fig:seedIllustration} for a sample $T_2$ slice. Here $r_f$ corresponds to the radius of the red region  selected as  foreground, and $r_b$ refers  to the background  brain region having green boundary. The bounding box is  drawn in yellow in the figure.

\section{Experimental Results}
\label{sec:setup}
The CADe system was modeled on the  BRATS 2015 dataset \cite{BRATS2015}, consisting  of 220  patients with  High Grade Glioma (HGG) over 155 slices from the four  MRI modalities $T_1$, $T_{1c}$, $T_2$, and $T_{2FLAIR}$,  along with their segmented ``ground truth"  about four intra-tumoral classes,  viz. edema, enhancing tumor, non-enhancing tumor, and necrosis. The data was aligned as $T_{1c}$, skull stripped, and interpolated to $1~mm^3$ voxel resolution. The total slice count for the entire dataset was $34,100$, with  each  slice being of size $240 \times 240$. The slices were resized to $96 \times 96$ before  training on  $210$ samples and testing on the remaining $10,$  with  final bounding box  being interpolated back  to  the original input slice  dimension.  Training  phase of  {\scshape{c-cnn}}  consisted of labeling  slices as ``nomal'' or ``abnormal'',  based on the ground truth. In case of {\scshape{d-cnn}},  the model  generated  the rectangular bounding box  fully enclosing  the  tumor region (for ``abnormal'') and  encoded as   $\left\lbrace x_{ul}, y_{ul}, width, height\right\rbrace$.

The {\scshape{c-cnn}} and {\scshape{d-cnn}}  networks were developed using Theano \cite{Bergstra2010_theano}, with a wrapper library Keras \cite{Chollet2015} in Python. The experiments were performed on a Dell Precision 7810 Tower with 2x Intel Xeon E5-2600 v3, totalling 12 cores, 256GB RAM, and NVIDIA Quadro K6000 GPU with 12GB VRAM. The operating system was Ubuntu 14.04. Segmentation of tumor regions was performed using ITK-SNAP \cite{itksnap} software.

After classification and detection by the {\scshape{c-cnn}} and {\scshape{d-cnn}}, the bounding box was  used to select seeds for
subsequent segmentation by  Algorithm \ref{alg:genseeds}. This constitutes the Automated GrowCut (AGC) segmentation. A comparative study is also provided with a manual  initialization  from seeds, using ground truth about  the foreground and background regions.  This is termed Semi-Automated GrowCut (SGS) segmentation.

\subsection{Dectection}
The performance of the two networks was quantitatively evaluated using eqns. (\ref{eq:accuracy})-(\ref{eq:dsc}). The  {\scshape{c-cnn}}  achieved an accuracy of $94.25\%$, with an area under the ROC curve of $0.9825$. The precision and recall values  were observed to be $0.9451$ and $0.9507$, respectively. The $F_\beta$ score, with $\beta = 1$, was $0.9479$. The  high recall rate implies detection of a large number of abnormal slices, while the high precision demonstrates accurate distinction  between normal and abnormal slices. In case  of {\scshape{d-cnn}}, the MAE was  $3.12$ pixels with standard deviation of  $7.02$ while generating the  bounding box. The DSC measured the overlap to be $0.8631$.

\begin{table}
	\centering
	\caption{Comparative study of DSC for detection}
	\label{tab:comp}
	\begin{tabular}{ll}
		\toprule
		\textbf{Method} & \textbf{DSC} \\ \midrule
		\textbf{Proposed CADe}                            & \boldmath{ $0.8631$}                           \\
		Saha \emph{et al.} \cite{Saha2012_quick}   & ~$0.4635$                           \\ \bottomrule
	\end{tabular}
\end{table}

\begin{figure}
	\includegraphics[width=\linewidth]{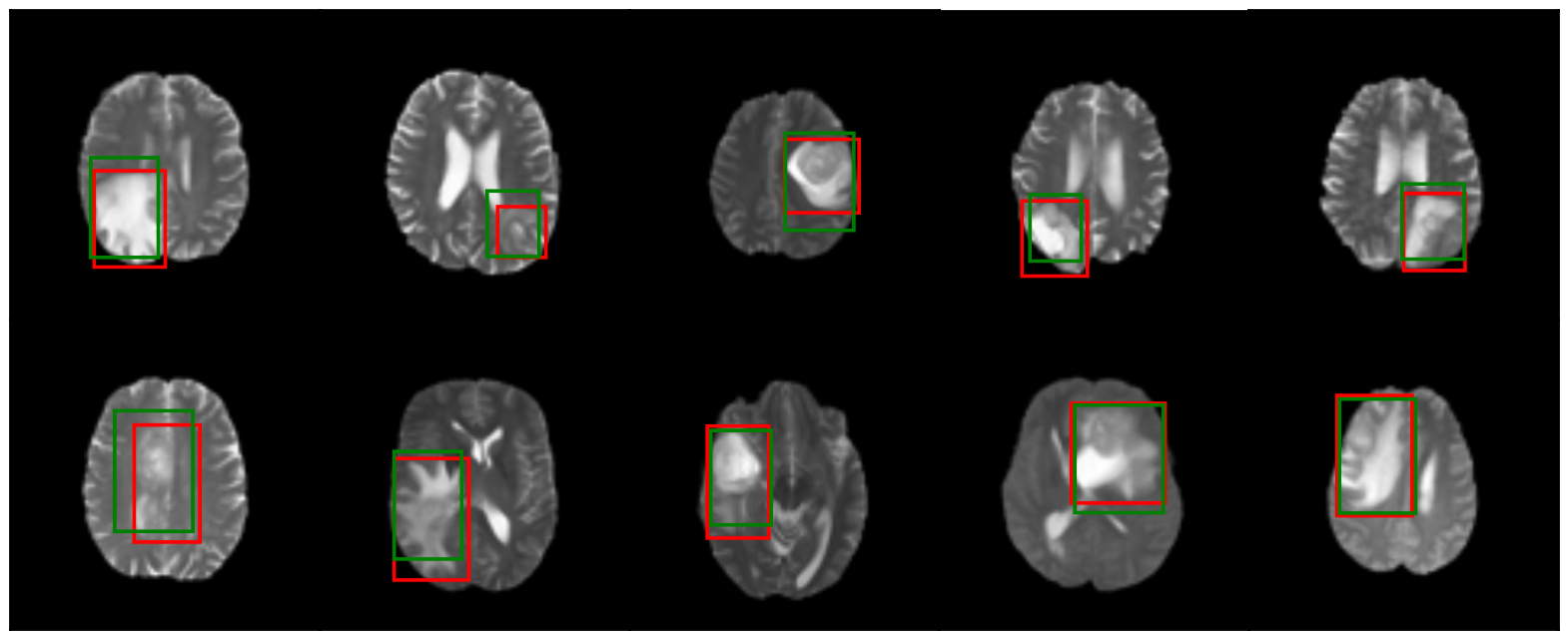}
	\caption{Bounding box on  10  sample patient slices $(T_2)$ generated by  the CADe system. Red rectangle denotes ground truth and green rectangle indicates  the predicted response.}
	\label{fig:qual-results}
\end{figure}

We also present a comparison of DSC, with that of the earlier approach by Saha \emph{et al.} \cite{Saha2012_quick}, in Table \ref{tab:comp}.
The overlap between the ground truth (target) and predicted regions (by bounding box) is found to be  higher in our proposed CADe system.
The qualitative result for the CADe system is presented in Fig. \ref{fig:qual-results}. The results demonstrate that the bounding boxes predicted by the {\scshape{d-cnn}}  closely resemble the original ground truth.

\begin{figure}
	\includegraphics[width=\linewidth]{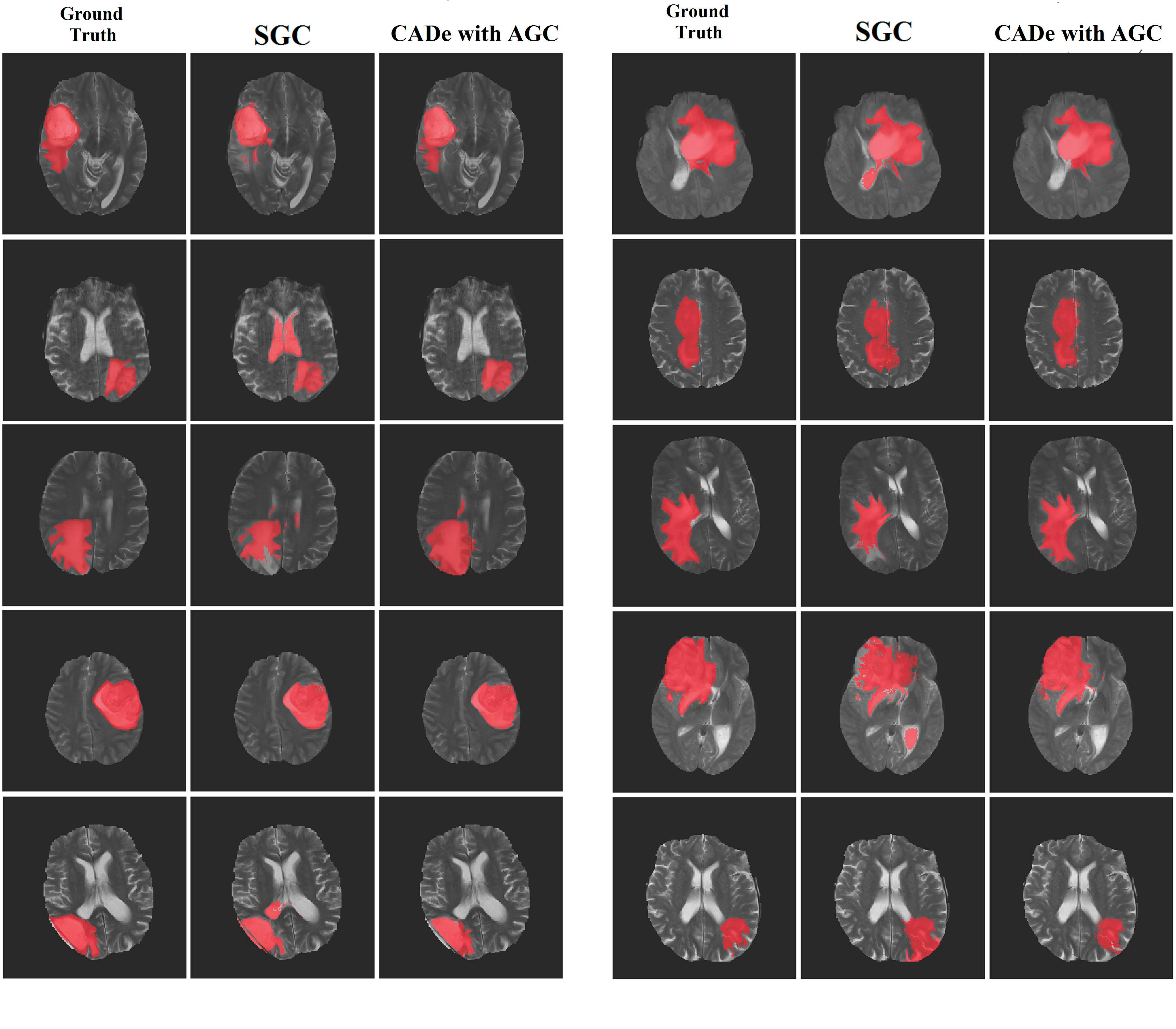}
	\caption{Comparative study of segmentation of ten sample patients}
	\label{fig:qual}
\end{figure}

\subsection{Segmentation}
Fig. \ref{fig:qual} presents a qualitative comparison (over 10 sample patients) of the segmentation obtained by the semi-automated SGC, involving manual insertion of seeds in the foreground and background, with that of our fully automated CADe system using AGC. It is clearly observed that the proposed model accurately simulates the ground truth.

Table \ref{tab:segsgc}  provides a quantitative comparison  between these algorithms, by computing the  mean and standard deviation (SD) over three runs, corresponding to  DSC for segmentation  for the $T_{2}$ and $T_{2FLAIR}$ sequences. The semi-automated SGC involves three independent observers to insert the seed points in the foreground and background regions. It is observed from  the table  that  inter-observer SD exists in SGC. On the other hand, the deterministic nature of  our automated CADe system enables complete elimination of any deviation over seed initialization. As SD $ \ll 10^{-5}$, over three runs, it was considered to be approximately zero and hence is  not reported in the table. The last row of the table presents the corresponding average DSC (for segmentation) over the 10 sample patients. It is evident that the automated AGC, used by our CADe system, provides an overall better match over both $T_2$ and $T_{2FLAIR}$ sequences.

\begin{table}
	\centering
	\setlength\tabcolsep{0.8mm}
	\caption{Comparative study of DSC [Mean(Standard Deviation)] for segmentation in 10 sample patients}
	\label{tab:segsgc}
	\begin{tabular}{C||CC||CC||}
		\toprule
		\textbf{Patient} &  \multicolumn{2}{c||}{\boldmath{$T_{2FLAIR}$}}      &  \multicolumn{2}{c||}{\boldmath{$T_2$}}       \\
		\textbf{ID} & SGC  &AGC  & SGC & AGC \\ \hline
		1& 0.72(0.03) & 0.81 &  0.78(0.00) & 0.82 \\
		2& 0.77(0.08) & 0.90 &  0.68(0.10) & 0.85 \\
		3& 0.72(0.05) & 0.89 &  0.79(0.00) & 0.84 \\
		4& 0.84(0.02) & 0.90 &  0.86(0.01) & 0.89 \\
		5& 0.55(0.04) & 0.78 &  0.61(0.01) & 0.72 \\
		6& 0.79(0.06) & 0.88 &  0.68(0.12) & 0.92 \\
		7& 0.62(0.10) & 0.74 &  0.57(0.12) & 0.74 \\
		8& 0.68(0.03) & 0.83 &  0.65(0.04) & 0.82 \\
		9& 0.76(0.07) & 0.90 &  0.68(0.04) & 0.89 \\
		10&0.60(0.11) & 0.72 &  0.52(0.12) & 0.68 \\ \hline
Average& 0.71(0.09) & 0.84(0.07) & 0.63(0.10) & 0.82(0.08) \\\bottomrule
	\end{tabular}
\end{table}

\subsection{Analysis of architecture}

 The proposed network design for {\scshape{c-cnn}} and {\scshape{d-cnn}} were next evaluated with respect to several variations  in architecture. Considering the architecture of Sec. \ref{baseline} as the ``baseline'' model, four experiments were performed as enumerated below.

\begin{enumerate}
 	\item[E1.]  Training without any data augmentation. Absence of data augmentation typically leads to overfitting,  with most artificial augmentation involving random rotations, width or height shifts, horizontal  or vertical  flipping, etc. \cite{Pereira2016}.
  	\item[E2.] Using larger than $3 \times 3$  kernels in the convolution layers.
    \item[E3.] Employing deeper layers (networks).
    \item[E4.] Exploring LeakyReLU \cite{Maas2013_lrelu} as layer-wise non-linearity, instead of standard ReLU.
\end{enumerate}

In order to establish the statistical significance of our baseline model, a pairwise t-test is performed between the corresponding DSC scores (with a null hypothesis that the models being compared are similar).
We set a  threshold of $0.05$, with the null hypothesis being rejected when  the computed $p$-value from a test between a pair of  models becomes  lower than this threshold. It implies that the difference between mean DSCs is likely to represent an actual difference between the pair of  models being compared.

\subsubsection{ \mbox{\scshape{c-cnn}}} Table \ref{tab:c-cnn_results} provides a  study of comparative classification performance  of the four variants over the baseline  {\scshape{c-cnn}} architecture. Quantitative evaluation is made in terms of accuracy,   Precision, Recall, $F_{\beta}$  of eqns. (\ref{eq:accuracy})-(\ref{eq:beta}), and Area Under the Curve (AUC).

\begin{table}
	\centering
	\caption{Comparative study of   {\scshape{c-cnn}} variants}
	\label{tab:c-cnn_results}
	\begin{tabular}{@{}C|CCCCC@{}}
		\toprule
		\textbf{Experiment} & \textbf{Accuracy} & \boldmath{\boldmath{AUC}} & \textbf{Precision} & \textbf{Recall} & \boldmath{F_{\beta}} \\ \midrule
		\textbf{Baseline}   & \boldmath{\boldmath{94.25}}\%  &0.9825                 & 0.9451             & \boldmath{\boldmath{0.9507}} & \boldmath{\boldmath{0.9479}}               \\
		\text{E1}                  & 93.87\%           & \boldmath{\boldmath{0.9836}}       & 0.9458             & 0.9424          & 0.9441               \\
		\text{E2}                  & 93.29\%           & 0.9817         		& 0.9441             & 0.9330          & 0.9386               \\
		\text{E3}                  & 92.32\%           & 0.9824         		& 0.9496    & 0.9084          & 0.9286               \\
		\text{E4}                  & 91.35\%           & 0.9798         		& \boldmath{\boldmath{0.9521}}             & 0.8873          & 0.9185               \\ \bottomrule
	\end{tabular}
\end{table}

It is observed from Table~\ref{tab:c-cnn_results} that E1 leads to overfitting, thereby causing a drop in detection Accuracy, Recall and $F_{\beta}$, with poor generalization. On the other hand, Precision over the training set was higher than the baseline model. Analyzing  Table \ref{tab:c-cnn_results}, we  observe that data augmentation improves  delineation between normal and abnormal tissues in {\scshape{c-cnn}}.

Employing larger fitter (or kernel) sizes in each convolution block (by E2) as compared to the $3 \times 3$ size the baseline model,
and  going up-to $7 \times 7$ increasing by $2$ units over each pair of convolution layer, resulted in  an increase of approximately 1.8 times in the number of tunable parameters. The higher network size  produced  increased computational overhead with higher training and testing times.
Examining Table \ref{tab:c-cnn_results} we note that having larger kernels leads to overfitting and degradation of generalization performance due to increased trainable parameters.

It has been consistently mentioned in deep learning literature that going deeper with convolutions may increase performance. The whole idea behind deep learning is to train as \textit{deep} networks as possible. Since a
conventional {\scshape{c-cnn}} employs pooling layers which reduce the dimension of its input, there appears an inherent upper limit to the depth  before the network exhausts itself of  input feature maps. In  experiment E3, we tested with three additional layers [a convolution block (two convolution layers) and a pooling layer] being  added just before the fully connected layers. The parameters of these  newly added   layers mimic the ones before them. The E3 version of {\scshape{c-cnn}},  with 15 layers, exhibited  poorer performance than the baseline model,   as observed from Table~\ref{tab:c-cnn_results}. However Precision on the training set was higher.

It is argued that imposing a strict condition to zero out the negative neuron activation, in ReLU,  may lead to gradient impairment and subsequent  adjustment of weights in the network. As a result a new variant called LeakyReLU \cite{Maas2013_lrelu},  with activation function $max(0,a) + \alpha~*~ min(0,a)$, where $\alpha$ is the leakiness parameter, was employed. The  function is designed to ``leak'' negative gradient instead of zeroing it \cite{Pereira2016}. Here we investigate the use of LeakyReLU, instead of standard ReLU, under E4  with $\alpha = 0.2$ (since higher values resulted in divergence of training). It is clear from Table~\ref{tab:c-cnn_results} that the generalization performance was poorer over the baseline model, while the Precision on training set was higher.

\subsubsection{{\scshape{d-cnn}}} Table~\ref{tab:d-cnn_results} presents a comparative analsysis of the four variants over the baseline {\scshape{d-cnn}}  architecture. Quantitative evaluation is provided in terms of MAE and DSC of eqns.~(\ref{eq:mae}) and (\ref{eq:dsc}). It is observed that absence of data augmentation lead to poorer generalization performance, as compared to our baseline model. On subjecting the baseline model  to $t$-test against  E1, a $p$-value = $0.0069$ demonstrated its  statistical significance.

\begin{table}
	\centering
	\caption{Comparative study of  {\scshape{d-cnn}} variants}
	\label{tab:d-cnn_results}
	\begin{tabular}{@{}C|CC@{}}
		\toprule
		\textbf{Experiment} & \textbf{MAE}    &  \textbf{DSC} \\ \midrule
		\textbf{Baseline}            & \boldmath{\boldmath{3.12}} \pm \boldmath{\boldmath{7.02}} & \boldmath{\boldmath{0.8631}}       \\
		E1                  & 3.53 \pm 6.93  & 0.8401       \\
		E2                  & 3.51 \pm 6.52  & 0.8606       \\
		E3                  & 4.56 \pm 8.83 & 0.8003       \\
		E4                  & 3.35 \pm 7.08 & 0.8429       \\ \bottomrule
	\end{tabular}
\end{table}

For large kernels, the increase  was upto $9 \times 9$ due to the extra convolution block. This resulted in  an increase of tunable parameters by around 3.3 times, with poorer performance in Table~\ref{tab:d-cnn_results}. A deeper architecture, by E3, generated a network  of 18 layers. However the generalization performance in Table~\ref{tab:d-cnn_results} was poorer than that of our baseline model. The larger size resulted in increased training and testing overheads. The pairwise $t$-test performed between the baseline {\scshape{d-cnn}} and E3, over DSC,  returned a  $p$-value = $6.4 \times 10^{-11}$,  demonstrating its  statistical significance. Hence it can be inferred  that going deeper with convolutions did not help improve the performance. Use of LeakyReLU in E4 resulted in poorer performance as well. Statistical significance of our baseline model was   proven by a $p$-value of $0.019$.

\section{Conclusions}
\label{sec:conclusion}
An automated Computer Aided Detection (CADe) system has been developed, using Convolution Neural Networks, for detecting
and segmenting high grade gliomas from brain MRI. The concept of bounding  box is employed to detect tumor cases, with subsequent localization of the  abnormality  from individual MR slices. Two ConvNet models {\scshape{d-cnn}} and {\scshape{d-cnn}} were designed for the purpose. The detection and delineation results on the BRATS 2015 database, demonstrated the effectiveness of the choices of hyperparameters was studied.  Comparative studies with related methods established the superiority of our CADe system.

\bibliographystyle{ieeetr}

\end{document}